\def\year{2021}\relax
\documentclass[letterpaper]{article} 
\usepackage{aaai21}  
\usepackage{times}  
\usepackage{helvet} 
\usepackage{courier}  
\usepackage[hyphens]{url}  
\usepackage{graphicx} 
\usepackage{natbib}  
\usepackage{caption} 
\usepackage{graphicx}  
\usepackage{amsfonts}
\usepackage{amsmath}
\usepackage{booktabs}
\usepackage{amssymb}

\usepackage[ruled,vlined]{algorithm2e}
\usepackage{verbatim}
\usepackage{bm}
\usepackage{multirow}
\usepackage[noend]{algpseudocode}
\title{Learning from My Friends:\\ Few-Shot Personalized Conversation Systems via Social Networks}

\author{

    Zhiliang Tian,\textsuperscript{\rm 1,4}\thanks{This work was partially done when Zhiliang Tian was
    an intern at Tencent AI Lab} Wei Bi\textsuperscript{\rm 2},\thanks{Corresponding author} Zihan Zhang,\textsuperscript{\rm 1} Dongkyu Lee,\textsuperscript{\rm 1} Yiping Song,\textsuperscript{\rm 3} Nevin L. Zhang\textsuperscript{\rm 1,4}
}
\affiliations{

    \textsuperscript{\rm 1}Department of Computer Science and Engineering, \\
The Hong Kong University of Science and Technology, Hong Kong SAR, China\\
    \textsuperscript{\rm 2}Tencent AI Lab, Shenzhen, China \\ 
    \textsuperscript{\rm 3}National University of Defense Technology, Changsha, China\\
    \textsuperscript{\rm 4}HKUST Xiao-i Robot Joint Lab, Hong Kong SAR, China\\

    \{ztianac,dleear,lzhang\}@cse.ust.hk victoriabi@tencent.com zzhangeo@connect.ust.hk songyiping@pku.edu.cn

}

\begin{document}

\maketitle
\begin{abstract}
    Personalized conversation models (PCMs) generate responses according to speaker preferences. Existing personalized conversation tasks typically require models to extract speaker preferences from user descriptions or their conversation histories, which are scarce for newcomers and inactive users. In this paper, we propose a few-shot personalized conversation task with an auxiliary social network. The task requires models to generate personalized responses for a speaker given a few conversations from the speaker and a social network. Existing methods are mainly designed to incorporate descriptions or conversation histories. Those methods can hardly model speakers with so few conversations or connections between speakers. To better cater for newcomers with few resources, we propose a personalized conversation model (PCM) that learns to adapt to new speakers as well as enabling new speakers to learn from resource-rich speakers. Particularly, based on a meta-learning based PCM, we propose a task aggregator (TA) to collect other speakers' information from the social network. The TA provides prior knowledge of the new speaker in its meta-learning. Experimental results show our methods outperform all baselines in appropriateness, diversity, and consistency with speakers.
\end{abstract}

\section*{Introduction}

Recently, there has been a boom in research on neural conversation models \cite{NRM_Shang15} due to the accessibility of vast conversational data on social media (e.g. Twitter). To generate appropriate and lively responses, researchers have proposed personalized conversation tasks that require models to customize responses for specific speakers, since different speakers tend to have different styles or preferences for their responses. There are two subtypes of such tasks. Description-conditioned tasks \cite{yang2017personalized,persona} require models to customize responses according to explicit speaker descriptions. These descriptions may come from human annotations \cite{persona} or user profiles in social media \cite{mazare2018training}. Speaker descriptions are not always available due to the cost of annotation and privacy concerns in social media. Conversation-conditioned tasks \cite{lijiwei, Exploring} require models to generate personalized responses by exploiting speakers' preferences from their conversation histories. In reality, conversation histories may provide very few utterances of a particular speaker, which makes it hard to capture speaker preferences, especially for the newcomers or inactive users.

To better cater for newcomers, we propose a few-shot personalized conversation task with an auxiliary social network. The task has three characteristics: 1. During training, models cannot access information about the speakers in the testing set (i.e. newcomers); 2. During testing, there are only a few samples available for each speaker, which are collectively referred to as the support set; 3. There is a social network among all the speakers. Given the input query, our task requires a conversation model to generate a response for a new speaker with the help of the social network and a few (i.e. 10) past conversation samples from the speaker. 


It is difficult to characterize the preferences of a new speaker from only a few conversation samples. Social networks can help here. In a social network, neighbors are users who follow each other, and they usually share interests and have similar chatting preferences. As our observation on our dataset, on average, the response similarity between two neighbors (0.47) is higher than that between two random speakers (0.38). \footnote{The gap between 0.47 and 0.38 is quite large in this evaluation metric, as a contrast, the gap between  similarity of two responses from one speaker (0.50) and 0.47 is only 0.03.} Consequently, we can utilize conversation histories of neighbors to help to determine preferences of a newcomer. In this way, we can handle newcomers even when there are no descriptions or a few conversations available.

Existing conversation-conditioned PCMs can be applied to our proposed task. \citealt{lijiwei} employ speaker embedding to capture speaker preferences. Based on this, \citealt{emnlp19_VHUCM} pre-built a conversation graph to describe speaker relations and learn node2vec embeddings \cite{node2vec} over the graph. The node2vec embeddings serve as the initial speaker embeddings in conversation models. To adopt such methods to our setting, we first train the models on the large scale training set; and then fine-tune them on a few conversations from the target speaker. However, training speaker embeddings or building a reliable conversation graph requires many samples from the target speaker, which is not always available in our setting. Model-agnostic meta-learning (MAML) \cite{maml_2017icml} based PCMs \cite{paml} is capable of fast adapting to new speakers by learning the adaptation ability across speakers. However, it aims to be a good initial model that is satisfactory for all speakers' adaption, instead of caring about the speakers' characteristics and relations. Due to the weakness in modeling speaker relations and distinguishing characteristics of speakers, this paradigm is not effective at making use of other resource-rich speakers.

In this paper, we propose a PCM that learns to adapt to new speakers as well as enables new speakers to leverage information from resource-rich speakers via a social network. Specifically, we first construct a MAML based PCM, where each speaker functions as a task. Then, we propose a task aggregator (TA) to collect resource-rich speakers' information according to speaker relations in the social network. The task aggregator explicitly represents each task, and the task representation serves as a task prior for the new speaker in its meta-learning.

In this way, our model utilizes the information of resource-rich speakers to augment low-resource speakers, and hence addresses the data deficiency issue on low-resource speakers. The experimental results show that our methods outperforms all alternative methods in terms of appropriateness, informativeness, and consistency.


Our contributions are summarized as follows,
\begin{itemize}
    \item We propose a few-shot personalized conversation task with a social network that better caters for new speakers.
    \item We design some variants of existing PCMs to our proposed task as strong baselines for our task.
    \item We propose a novel method that learns to learn from resource-rich speakers to customize responses for new speakers, and our model surpasses all the baselines.
\end{itemize}

\section*{Related Work}
\subsection*{Conversation Models}
Large scale conversation corpora from social media lead to a great success of retrieval-based methods \cite{yan2016learning,zhou2016retrieve} and generation-based methods \cite{seq2seq,NRM_Shang15,zhao2017CVAE}. To enhance the conversation models, \citealt{aaai16_hred,tian2017acl} proposed context-aware conversations that consider the previous utterances in current dialogue session. To ensure the topic consistency between the input and output, researchers proposed topic-aware models to capture and emphasize input's topic \cite{xing2017topic,peng2019topic}. Some papers utilized the background knowledge \cite{moghe2018towards,CMRACL2019,ren2019thinking} and commonsense knowledge \cite{zhou2018commonsense,Young2018Augmenting,MyACL2019} in conversations.

Researchers proposed to capture speakers' temporal status to generate appropriate responses. Emotion-aware dialogue models detect speaker's feelings and make appropriate emotional responses \cite{zhou2018emotional,fung2018empathetic,saha2020towards}. Multi-party dialog \cite{multiparty_meng2018LREC,multiparty_zhang2018aaai,multiparty_ruiyan_hu2019gsn} distinguished different roles of different speakers within a dialogue session.

\subsection*{Personalized Conversation Models}
Personalized conversation models capture speaker's characteristic, such as language behaviors, styles, and hobbies. \cite{lijiwei,persona,olabiyi2018persona,cmaml_acl2020song,user_attribute_AAAI2020pre}. Some researchers focused on the conversational agent being aware of the human’s personality and adjusting the dialogue accordingly \cite{persona_aware_2009,persona_aware_2017,persona_aware_2019}. Others assigned the personality to the conversational agent and encouraged the agent to generate personalized responses. Our task falls into the latter category, which consists of two types: description-conditioned and conversation-conditioned.

Description-conditioned personalized conversation tasks capture the speaker preference by explicit speaker description. \citealt{persona} described the speaker with a few profile sentences and built Persona-Chat dataset by crowd-sourcing. \citealt{mazare2018training} summarized speaker comments from Reddit into several sentences as speaker descriptions. \citealt{user_attribute_Qian2018Assigning,user_attribute_2019_persona_dialog} defined some profile key-value pairs, such as gender, age, and location. As for the methods on the above tasks, \citealt{yang2017personalized} fine-tuned a speaker-specific model based on a pre-trained speaker-independent model. \citealt{persona} encoded persona description into a memory network and employed a seq2seq model to generate responses with the memory, and \citealt{persona_memorynet_emnlp18} refined the memory network with attention mechanism \cite{attention}. \citealt{yavuz2019deepcopy} explored the use of copy mechanism in personalized dialogue models. \citealt{user_attribute_AAAI2020pre} imported a pre-trained GPT to transformer-based personalized dialog models.
Different from this type of task, our task does not require the external description of speakers.

Conversation-conditioned personalized conversation tasks describe speakers with their conversation histories. \citealt{lijiwei}, the first to propose neural personalized conversation models, implicitly modeled speakers' personas using speaker embeddings. \citealt{Exploring} modeled the speaker personality together with conversational contexts by a hierarchical recurrent encoder-decoder model \cite{aaai16_hred} with speaker embeddings. \citealt{emnlp19_VHUCM} learned node2vec \cite{node2vec} embeddings from a conversation graph to initiate the speaker embeddings. The above three papers cannot handle new speakers, and they require many dialogues for each speaker. Our task is designed to handle new speakers with very few resources. \citealt{paml} learned to adapt to new speakers with few shots, but their framework is not effective at incorporating and utilizing speaker relations. Our model caters for new speakers and encourages resource-rich speakers to help low-resource speakers in a social network.


\section*{Task Definition}
Our task aims to enable conversation systems to act as a given speaker and make personalized responses to the input query according to the speaker preference. Our task requires the systems to follow a speaker preference or style, even if the speaker is a newcomer with little information.

Our setting consists of a training set, a testing set, and a validation set. Each of those sets consists of speakers $\mathbb{S}=\{s_1,s_2...,s_n\}$, the conversation samples $D_{s}=\{\langle q,r_{s}\rangle_1, \langle q,r_{s}\rangle_2, ..., \langle q,r_{s}\rangle_m\}$ involving speaker $s$, and a graph $\mathcal{G}$ that represents the social network. The set is denoted as $\{\mathbb{S}, \bigcup_{s \in \mathbb{S}}  D_{s}, \mathcal{G}\}$. A conversation sample $\langle q,r_{s}\rangle$ is a pair of input queries and the corresponding responses made by speaker $s$. In the undirected graph $\mathcal{G}$, speakers act as the nodes, relations between speakers act as the edges, and two nodes bridge an edge once two speakers follow one another.

The speakers in the testing or validation set are newcomers, and they do not overlap with speakers in the training set. The graph in the training set is composed of only the training speakers, and the graph in testing or validation contains all speakers. For each newcomer $s$, we can access only $K$ conversation samples before inference (prediction), which make up the support set $D^{sup}_{s}$. We are required to make inference on other $K$ conversation samples from the speaker, which make up the query set $D^{qu}_{s}$. $K$ is the shot number for few-shot learning. In summary, our task is to generate a response given a query $q$, a current speaker $s$, the speaker's support set $D^{sup}_{s}$, and a social network $\mathcal{G}$.

\section*{Methodology}
\subsection*{Architecture}
\begin{figure}[htb]
	\centering
	\includegraphics[width=.43\textwidth]{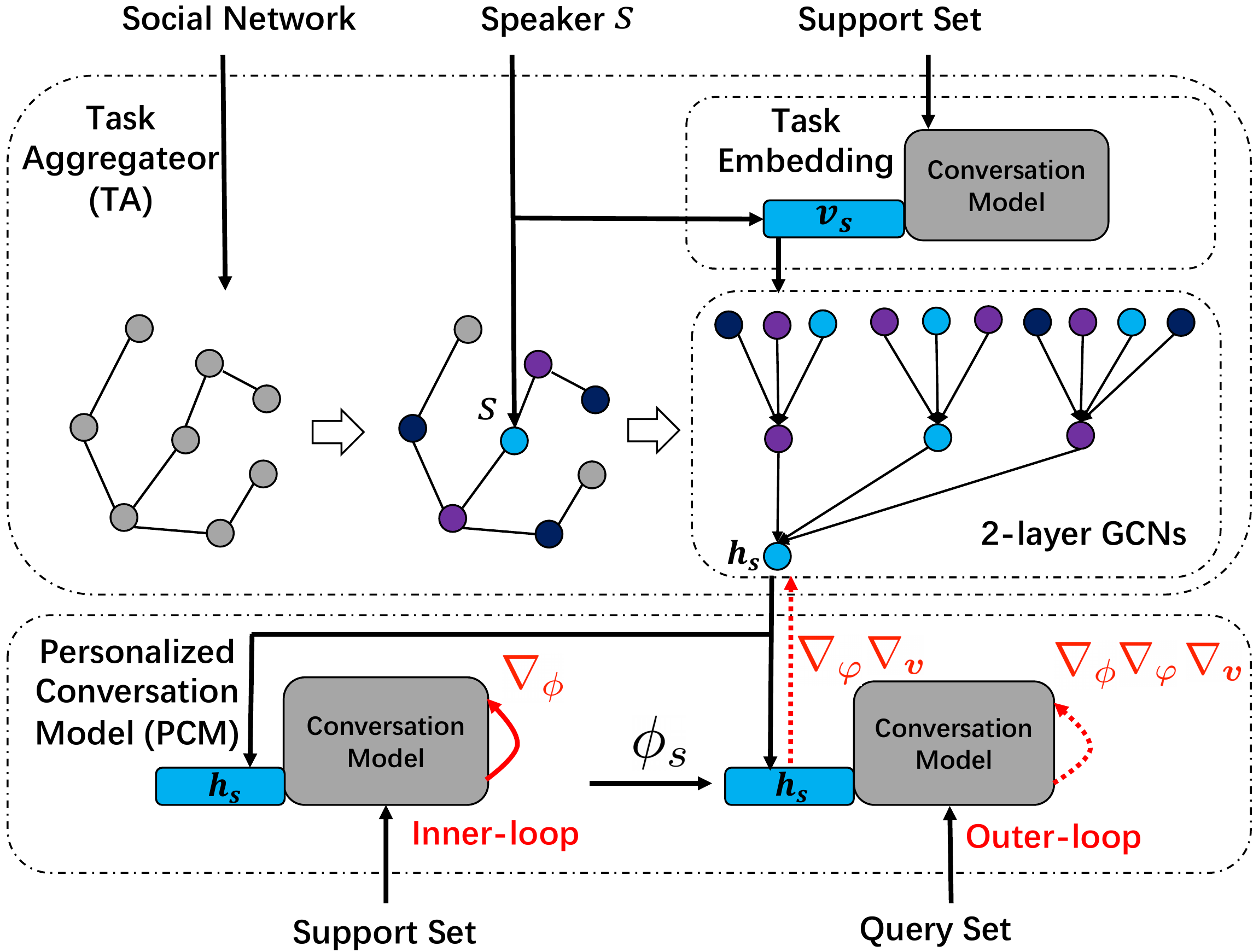}
	\caption{The structure of our model. Solid and dashed red arrows indicate the backward pass of PCM's inner-loop and outer-loop, respectively. Blue, purple, and mazarine blue blocks represent variables of speaker $s$, its 1-hop neighbors, and its 2-hop neighbors, respectively. The whole figure shows the training mode; in the testing mode, operations shown by dashed red arrows do not work. In our full model, all components in the figure works; in our variant of \textit{Ours$-$SelfEmb}, the task embedding block does not work. Eq.~\ref{eq:MAML_step3} to \ref{eq:update_phi_carphi_pi_3} show the optimization of $\phi$, $\varphi$, and $\bm{v}$ in the inner- and outer-loops.}
	\label{fig:train}
\end{figure}

Our model consists of two sub-modules as shown in Fig.~\ref{fig:train}: 
\begin{itemize}
    \item \textit{Personalized Conversation Model (PCM)} makes personalized responses according to speaker preferences. A transformer-based speaker-independent conversation model (CM) acts as the base model of PCM, denoted as $f_{\phi}^{CM}$ with its parameters $\phi$. PCM employs MAML to learn to adapt CM to new tasks (i.e. speakers) with few examples so as to make personalized responses.
    \item \textit{Task Aggregator (TA)} learns to study from related tasks (i.e. speakers) to assist the target task. Based on PCM, The TA explicitly represents tasks with embeddings, and then learns to aggregate embeddings from other tasks to refine the representation of the target task. The refined representation serves as the prior knowledge of the target task in PCM. We denote the operation of obtaining task embeddings as $f^{Emb}_{\bm{v}}$ with its parameters $\bm{v}$, and operation of aggregating related task embeddings as $f_{\varphi}^{AG}$ with parameters $\varphi$.
\end{itemize}

The TA provides the prior knowledge of the target task for PCM, while PCM provides the feedback for the TA's back-propagation. The TA regards each task as a training sample, and PCM's performance on an entire task serves as the feedback for a TA's training step. We introduce the two sub-modules and our training and testing methods in the following subsections.

\subsection*{Personalized Conversation Model (PCM)}
\label{sec:pcm}
PCM learns to generate personalized responses for new speakers via MAML \cite{maml_2017icml}, where it treats new speakers as new tasks. First, PCM constructs a speaker-independent conversation model (CM). The CM is a conventional transformer model \cite{transformer} with the ability to generate response $r$ to the input query $q$. We formulize the CM as $r=f_{\phi}^{CM}(q)$. 
Then, we apply MAML to the CM to build a personalized model. MAML can learn good initial parameters for CM that achieves rapid adaptation to new speakers with a few examples. For each training step, we sample a batch of speakers with their samples $\{S=\{s_1,...,s_{l}\}, \bigcup_{s \in S} D_{s}\}$, and then build a support set $D^{sup}_{s}$ and query set $D^{qu}_{s}$ with $K$ distinct examples sampled from $D_{s}$, respectively. 


The training of MAML consists of an inner-loop phase and an outer-loop phase. In the inner-loop, MAML trains the conversation model $f_{\phi}^{CM}$ on $D^{sup}_{s}$ and adapts $f_{\phi}^{CM}$ to the speaker $s$ by updating its parameters from $\phi$ to $\phi_{s}$. In the outer-loop, MAML updates the parameters $\phi$ according to the performance of the model trained by the inner-loop phase, which is evaluated on the query set. Particularly, in the inner-loop, the parameters $\phi$ is optimized by the gradient of $f_{\phi_{s}}^{CM}$ with respect to $\phi$ across tasks in $S$. 

\subsection*{Task Aggregator (TA)}
While PCM customizes responses for the given single task, TA models task representations and relations among tasks with the purpose of providing the prior knowledge of a task for PCM. TA obtains conversation-conditioned task embeddings by $f^{Emb}_{\bm{v}}$, refines task embeddings with the social network and a task embedding aggregator $f^{AG}_\varphi$, and then feeds the refined task embeddings to PCM. TA is optimized according to  PCM's feedback; meanwhile, the task embeddings are also constrained by the graph structure of the social network.

\subsubsection{Conversation-conditioned Task Embeddings} We employ a task embedding layer, assign each task with an embedding, and learn the task embedding with conversations under this task. As traditional embedding layers, task embeddings start from randomly initialized parameters. During training, we fetch the task embedding $\bm{v}_{s}$ of the current task $s$, and combine the embedding with the conversation model (CM) by appending the embedding in front of the sequence of CM's input word embeddings. \footnote{We did try some different combinations of extra embeddings with the transformer model and picked the best one, and similar usage can be found in \cite{ctrl2019arkiv_keskar}.}
Afterwards, we train the combined model on the conversations from the support set $D^{sup}_{s}$ of the current task $s$. After the training, the task embedding $\bm{v}_{s}$ serves as the output of this component. We formulize this component as $\bm{v}_{s} = f^{Emb}_{\bm{v}}(s, D^{sup}_{s})$, where the inputs are current speaker $s$ and its support set, and $\bm{v}$ denotes the task embedding parameters.

Task embedding parameters $\bm{v}$ and CM's parameters $\phi$ are optimized iteratively. In the training of this component, we fix $\phi$ and optimize the embedding $\bm{v}$; in the training of PCM and we optimize CM's parameters $\phi$. (Sec.~\ref{sec:train}).


\label{sec:gcn}
\subsubsection{Task Embedding Aggregator via GCN} As the scale of $D^{sup}_{s}$ for new speakers is still too small for the task embedding learning, we enhance the task embedding by related resource-rich tasks.

Social networks naturally reveal the relationship between speakers. We construct an undirected graph $\mathcal{G}$, where speakers act as nodes, and two nodes bridge an edge once two speakers follow one another. Since we observed that neighbors (two speakers follow each other) have similar preferences in responding, we utilize task embeddings of neighboring speakers seen in the training set.

We employ 2-layer GCNs \cite{gcn2017iclr} to refine task embeddings with neighbors' embeddings. The original GCN \cite{gcn2017iclr} conducts the propagation on the whole graph. It is time-consuming to re-train with the whole graph for the newcomers, so we follow the neighbor aggregation variant of GCN \cite{GCNSurvey2019iclr}. We formulize this aggregator as $\bm{h}_{s} = f^{AG}_\varphi(\bm{v}_{s}, \mathcal{G})$. 



We also propose another model variant that does not acquire the conversation-conditioned task embedding $\bm{v}_{s}$ of the target task and relies only on the neighbors. That variant gets the final task representation $\bm{h}_{s}$ by feeding all the neighbor's embedding into the GCN without using $\bm{v}_{s}$. We discuss it in the experiment section with the notation of \textit{Ours$-$SelfEmb}.

\begin{algorithm}[htb]
    \IncMargin{2em}
    \caption{Training Algorithm}
    \label{al:training_algorithm}
    \KwIn{ $\{\mathbb{S}, \bigcup_{s \in \mathbb{S}}  D_{s}\}$: training set \\
    \qquad \quad $\alpha, \beta$: learning rates\\
    \qquad \quad $\lambda$: weight for negative sampling loss \\
    \qquad \quad $\mathcal{G}$: social network on training speakers \\
    }
    \KwOut{$\{\phi, \varphi, \bm{v}\}$: trained model parameters}
    
    Randomly initialize $\phi$, $\varphi$, $\bm{v}$. \\
    \While{not done}
    {
    Sample a batch of speakers with their samples 
    $\{S, \bigcup_{s \in S} D_{s}\} \sim \{\mathbb{S}, \bigcup_{s \in \mathbb{S}}  D_{s}\}$ \\
    \For{all $s$ in $S$} 
    {
    // Inner-loop (train the base model) \\
    Build $D^{sup}_{s}$ and $D^{qu}_{s}$ from $D_{s}$ as Sec.~\ref{sec:pcm}\\ 
    Get task embeddings $\bm{v}_{s}$ by $f^{Emb}_{\bm{v}}$ \\
    Get aggregated task embeddings $\bm{h}_{s}$ via GCN \\ 
    Evaluate $\nabla_{\phi} \mathcal{L}_{D^{sup}_{s}} (f^{CM}_\phi (\bm{h}_{s}, q))$ with respect to $K$ examples. \\
    Compute adapted parameters $\phi_{s}$ with gradient descent as Eq.~\ref{eq:MAML_step3} \\
    }
    // Outer-loop (meta-optimization across tasks) \\
    Evaluate $\nabla_{\{\phi,\varphi,\bm{v}\}} \sum_{s \sim S} \mathcal{L}_{D^{qu}_{s}} (f^{CM}_{\phi_{s}} (f^{AG}_\varphi(f^{Emb}_{\bm{v}}(\cdot)))$ with respect to $K$ examples. \\
    Evaluate $\nabla_{\{\varphi,\bm{v}\}} \lambda \sum_{s \sim S} \mathcal{L}^{ns}_{s} (f^{AG}_\varphi(f^{Emb}_{\bm{v}}(\cdot))$ with respect to $\mathcal{G}$. \\
    Update $\{\phi, \varphi, \bm{v}\}$ using Eq.~\ref{eq:update_phi_carphi_pi_1},~\ref{eq:update_phi_carphi_pi_2}, and \ref{eq:update_phi_carphi_pi_3}. \\
    }
\end{algorithm}

\subsubsection{Graph Structure Constraint via Negative Sampling}
$f^{Emb}_{\bm{v}}$ represents a task according to conversations. We argue that task representations should also be constrained by the graph $\mathcal{G}$, which reveals task relations. As neighbors share similar preferences, they ought to have similar task embeddings. Inspired by node2vec \cite{node2vec}, we employ a negative sampling loss \cite{word2vec2013nips} that forces task embeddings from two neighbors to be more similar than those from two randomly selected speakers as,
\begin{equation}
\label{eq:n2v_loss}
\mathcal{L}^{ns}_{s}=-(\sum_{i \in \mathcal{N}_{s}} \log \sigma (\bm{v}^{T}_{i} \bm{h}_{s}) + \sum^{K_{ns}}_{\substack{j \sim \mathbb{S} \\ k=1 }} \log \sigma (-\bm{v}^{T}_{j} \bm{h}_{s}))
\end{equation}

\noindent
where $\bm{h}_{s}$ comes from the aggregator, $K_{ns}$ denotes the number of negative samples, $\sigma$ denotes the sigmoid function, and $\mathbb{S}$ denotes the set of all speakers. Notice that the above two algorithms involving the graph are friendly to newcomers since their training only operates on the newcomers and their neighbors instead of conducting the propagation over the entire graph. 

\subsection*{Training and Testing}
\label{sec:train}
\subsubsection{Model Training.} Our whole model is the combination of the PCM and TA. 
We extend the PCM's training paradigm mentioned in Sec.~\ref{sec:pcm} to the whole model. 
We train the PCM via MAML as in Sec.~\ref{sec:pcm}, and train the TA in MAML's outer-loop. Algorithm~\ref{al:training_algorithm} and Fig.~\ref{fig:train} show the training procedure.

During the training, the TA first obtains the task embedding $\bm{h}_{s}$, and then feeds it to the PCM. The PCM training consists of inner-loop and outer-loop training. In the inner-loop, the PCM uses $\bm{h}_{s}$ in its feedforward but does not update the TA's parameters. In the outer-loop, the PCM uses $\bm{h}_{s}$ again and updates the parameters of both the PCM and TA according to the performance of the whole model on the query set. The optimization of the inner-loop and outer-loop is shown in Eq.~\ref{eq:MAML_step3} to~\ref{eq:update_phi_carphi_pi_3}. In both inner- and outer-loop training, PCM uses $\bm{h}_{s}$ by appending it in front of the sequence of CM's input word embeddings, where the formula of CM defined in Sec.~\ref{sec:pcm} turns to $r=f^{CM}_{\phi}(\bm{h}_s, q)$.

TA is to model task priors and relations among tasks, so we feed each task as a training sample to the TA. Then, we train and optimize the TA on a set of tasks. Since inner-loop trains samples within a task and outer-loop optimizes across tasks, we fix the TA's parameters in the inner-loop and optimize them in the outer-loop. In the outer-loop, the TA receives feedback about whether the current TA with the PCM performs well on each task and is optimized accordingly.

\begin{eqnarray}
    \label{eq:MAML_step3}
    \! \phi_{s} = \phi - \alpha \nabla_{\phi} \mathcal{L}_{(q,r) \sim D^{sup}_{s}} (f^{CM}_\phi (\bm{h}_{s}, q)).
\end{eqnarray}
\begin{eqnarray}
    \label{eq:update_phi_carphi_pi_1} 
    &\phi & \leftarrow \phi - \beta \nabla_{\phi} \sum_{s \sim S} \mathcal{L}_{D^{qu}_{s}}(f^{CM}_{\phi_{s}} (\bm{h}_{s}, q)),  \\ 
    \label{eq:update_phi_carphi_pi_2} 
    &\varphi & \leftarrow \varphi - \beta \nabla_{\varphi} \sum_{s \sim S}( \lambda \mathcal{L}^{ns}_{s} \\ & & \qquad
    + \mathcal{L}_{D^{qu}_{s}}(f^{CM}_{\phi_{s}} (f^{AG}_{\varphi} (\bm{v}_{s}, \mathcal{G}),q))), \nonumber \\
    \label{eq:update_phi_carphi_pi_3} 
    &\bm{v} & \leftarrow \bm{v} - \beta \nabla_{\bm{v}} \sum_{s \sim S} (\lambda \mathcal{L}^{ns}_{s} \\ & & \qquad + \mathcal{L}_{D^{qu}_{s}}(f^{CM}_{\phi_{s}} (f^{AG}_{\varphi} (f^{Emb}_{\bm{v}} (s, D^{sup}_{s}), \mathcal{G}),q))). \nonumber
\end{eqnarray}


\subsubsection{Model Testing.} Model testing adopts the training procedure except for the back-propagation in the outer-loop. The TA obtains the task representation $\bm{h}_{s}$ for the PCM. The PCM incorporates $\bm{h}_{s}$, updates its $\phi$ to $\phi_{s}$ in the inner-loop, and generates the final responses with the feedforward of the outer-loop. The only difference between training and testing is that testing does not need not to optimize parameters in the outer-loop (No dashed red arrows in Fig~\ref{fig:train} for testing).

\section*{Experimental Setting}
\subsection*{Dataset}
We collect the dataset from Weibo, an online chatting forum with social networks. We use 28.9K speakers with 2.02M samples for training, 1K speakers with 20K samples for testing, and 0.5K speakers with 10K samples for validation. We release the code of dataset construction \footnote{github.com/tianzhiliang/FewShotPersonaConvData}. 

\subsection*{Comparing Methods}

\begin{itemize}
\item \textbf{Base Models.}
We use two conventional conversation models, \textit{Seq2Seq} and \textit{Transformer}. We use the transformer as our base conversation model for all the following methods as it performs better than the seq2seq.

\item \textbf{Fine-tune.}
We fine-tune the transformer on the support set of the target speaker to obtain personalized models, noted as \textit{Transformer+F}. We increase the hidden dimension of the transformer to keep the same parameter scale with the following comparing models, noted as \textit{{{Transformer}$^{+}$}+F}. We employ MAML~\cite{maml_2017icml} to train the transformer and note it as \textit{PAML}~\cite{paml}.

\item \textbf{Fine-tune+Social Network.} \citealt{lijiwei}  encode speaker preferences with speaker embeddings. \citealt{emnlp19_VHUCM} pre-train node2vec embeddings over a speaker graph as the initial speaker embeddings. Since new speakers are unseen during training, their methods cannot obtain new speakers' embeddings. We adapt their original methods to our task by aggregating its neighbors’ embeddings and then fine-tune the embeddings on the support set. We denote them as \textit{Speaker+F+SN} and \textit{VHUCM+F+SN}, respectively.

\item \textbf{Ours.}
\textit{Ours} denotes our full model, and \textit{Ours$-$SelfEmb} is our model without using the conversation-conditioned embedding of the target speaker. 

\end{itemize}

\begin{table*}[htb]
\centering
\small
\begin{tabular}{c|c|c|c|c|c|c|c}
\hline
\multirow{2}{*}{} & \multicolumn{3}{c|}{Appropriateness}                                & Diversity                        & \multicolumn{2}{c|}{Consistency} & \multirow{2}{*}{$|$Param$|$} \\ \cline{2-7}
                  & Bleu1 Bleu2 Bleu3 Bleu4           & NIST           & CIDEr          & Dist1 Dist2         & Grd-F1          & TokSim         &                        \\ \hline
Seq2Seq           & 7.325 3.912 3.129 2.736           & 0.329          & 0.149          & 0.133 0.370         & 0.012           & 0.041          & 62M                    \\ 
Transformer               & 10.951 5.852 4.407 3.803          & 0.803          & 0.205           & 0.155 0.454           & 0.016           & 0.032          & 65M                    \\ \hline 
Transformer+F            & 11.109 5.902 4.466 3.459          & 0.790          & 0.182              & 0.157 0.485      & 0.020           & 0.052          & 65M                    \\ 
Transformer$^+$+F       & \textbf{12.219} 5.992 4.486 3.643 & 0.880          & 0.189          & 0.160 0.488           & 0.020           & 0.048          & 80M                    \\ 
PAML              & 11.200 6.192 4.691 4.012          & 0.884          & 0.201         & 0.174 0.495        & 0.021           & 0.054          & 80M                    \\ \hline 
Speaker+F+SN     & 11.670 6.182 4.615 3.922          & 0.927          & 0.176          & 0.133 0.427          & 0.018           & 0.035          & 80M                    \\ 
VHUCM+F+SN    & 10.981 5.461 3.976 3.357          & 0.892          & 0.141          & 0.142 0.460          & 0.018           & 0.043          & 80M                    \\ \hline
Ours$-$SelfEmb      & 11.815 6.277 4.676 4.010          & \textbf{0.971} & 0.196          & \textbf{0.183 0.586 } & 0.023           & 0.061          & 80M                    \\ 
Ours & 11.638 \textbf{6.315 4.803 4.145}          & 0.940          & \textbf{0.238} & 0.169 0.530          & \textbf{0.024}  & \textbf{0.062} & 80M                    \\ \hline
\end{tabular}
\caption{The overall performance on automatic evaluations.}
\label{tb:mainexp}
\end{table*}

\subsection*{Implementation Details}
\textit{Seq2Seq} follows \citealt{song2018ensemble} where the embedding and hidden dimensions are 620 and 1000. For the transformer-based model, we implement it as the original one \cite{transformer}, where the model dimension is 512, the stacked layer number is 6, and the head number is 8. We increase the model dimension to 640 on \textit{{{Transformer}$^{+}$}+F} for fair comparison on the parameter scale. Following \cite{paml}, in \textit{PAML} and ours, we used SGD for the inner-loop and Adam for the outer-loop with learning rate $\alpha=$ 0.01 and $\beta=$ 0.0003, respectively. For all methods, the batch size in training is 128. The vocabulary contains top 50k frequent tokens, and the maximum length of input queries and responses is 80. For training, we set $\lambda$ as 1 for the negative sampling. For inference, we apply a top-$k$ sampling decoding~\cite{edunov2018understanding} with $k$=5. The sizes of the support set and query set for testing users are 10 (10-shot).

\begin{table*}[htb]
\centering
\small
\begin{tabular}{c|c|c|c|c|c|c}
\hline
\multirow{2}{*}{} & \multicolumn{3}{c|}{Appropriateness}                                & Diversity                        & \multicolumn{2}{c}{Consistency} \\ \cline{2-7} 
                  & Bleu1 Bleu2 Bleu3 Bleu4           & NIST           & CIDEr          & Dist1 Dist2         & Grd-F1          & TokSim         \\ \hline
Ours & \textbf{11.638} 6.315 \textbf{4.803 4.145} & 0.940 & \textbf{0.238} & \textbf{0.169} 0.530  & \textbf{0.024}  & \textbf{0.062} \\ \hline
$-$NS Loss         & 11.002 5.784 4.343 3.730          & \textbf{0.954}          & 0.188          & 0.167 \textbf{0.550}          & 0.022           & 0.053          \\ 
$-$GCN             & 11.176 \textbf{6.959} 4.425 3.782          & 0.870          & 0.194          & 0.158 0.526          & 0.022           & 0.053          \\ 
$-$NS Loss $-$GCN     & 10.877 5.338 4.002 3.439          & 0.615          & 0.194          & 0.167 0.510           & 0.020           & 0.052          \\ 
$-$NS Loss $-$GCN $+$ RAG     & 10.989 5.771 4.202 3.552 & 0.779          & 0.192          & 0.175 0.491           & 0.020           & 0.048          \\ 

TA as MAML's base & 10.642 5.359 2.926 3.342          & 0.790          & 0.169          & 0.165 0.548          & 0.021           & 0.055          \\ \hline
\end{tabular}
        \caption{The performance of the ablation study. Row 2 to row 4 indicates the full model without the negative sampling, without the GCN layer, or without both of them. \textit{$+$ RAG} indicates applying retrieval-augmented generation methods. \textit{TA as MAML's base} optimizes TA by treating it as a part of MAML's base model.}
        \label{tb:ablation}
        \end{table*}

\subsection*{Evaluation Metrics}
We evaluate all methods with both automatic and human evaluations. For automatic evaluations, we evaluate the responses in three aspects:
\begin{enumerate}
    \item \textbf{Appropriateness}. 
    We evaluate the overall quality by measuring the matching between the ground-truth and generated responses on three metrics: \textit{BLEU} \cite{papineni2002bleu}, \textit{NIST} \cite{NIST_metric_doddington2002automatic}, and \textit{CIDEr} \cite{CIDEr_metric_CVPR2015}.
    \item \textbf{Diversity}. \textit{Dist-n} evaluates the proportion of n-grams of the generated responses~\cite{li2016diversitymmi,song2017diversifying}.
    \item \textbf{Consistency}. We measure the consistency between the generated responses and the speakers' conversation histories. \textit{TokSim} measures the consistency by token-level similarity with TF-IDF \cite{joachims1996probabilistic} weight and a stop-word filter. F1 grounding score \textit{Grd-F1} \cite{CMRACL2019,MyACL2020} is the harmonic mean of precision and recall, where precision and recall measures the ratio of retrieving a token from conversation histories as responses.
\end{enumerate}

For human evaluations, we hire five annotators from a commercial company to evaluate 300 samples randomly selected from 60 speakers. The annotators evaluate a 5-point scale on quality (\textit{H-Appr}), diversity (\textit{H-Div}), and consistency with speakers' previous conversations (\textit{H-Cons}). 

\section*{Experimental Results and Analysis}

\subsection*{Overall Performance}
\label{sec:exp1}
The results of all competing methods on automatic metrics are shown in Table~\ref{tb:mainexp}. Since \textit{Transformer} outperforms \textit{Seq2Seq}, we use \textit{Transformer} as the base generation model for all baselines. \textbf{Fine-tune} methods achieve the better performance than \textbf{Base models} methods, indicating that building personalized conversation models for each speaker helps to improve the overall performance. \textit{PAML} gets higher performance than other baselines, which demonstrates that meta-learning is more suitable to handle new speakers in our scenarios. In rows 4 to 9 of Table~\ref{tb:mainexp}, we compare all methods in the same scale of parameters (80M). Social network describes the relations among speakers, so the baseline methods of \textbf{Fine-tune+Social Network} are supposed to be better than \textbf{Fine-tune}. However, it is not the case in our experiments, which demonstrates that few conversations are inapplicable for training a speaker embedding, so the neighbors' information cannot contribute to the target speaker with the usages in \textbf{Fine-tune+Social Network}. 

\begin{table}[htb]
\small
\centering
    \begin{tabular}{c|c|c|c}
    \hline
                 & H-Appr                & H-Div                & H-Cons                \\ \hline
    Seq2Seq      &     2.77   &       1.84       &     2.27                 \\ 
    Transformer          &      2.98       &     \textbf{2.87}   &         2.67          \\ \hline
    Transformer+F       &         3.09              &    2.78       &        2.45      \\ 
   Transformer$^+$+F   &       3.15       &        2.72       &     2.88       \\ 
     PAML         &      3.05     &   2.68        &     3.02     \\ \hline
     Speaker+F+SN    &      2.91      &   2.66   &        3.02     \\ 
    VHUCM+F+SN   &     2.99    &    2.56       &     2.84  \\ \hline
   
    Ours$-$SelfEmb &    3.09        &    2.84  &     2.97      \\ 
    Ours         &     \textbf{3.17}   &   2.79     &   \textbf{3.09}    \\ \hline
    \end{tabular}
    \caption{The overall performance on human evaluations.}
    \label{tb:human}
    \end{table}

\textbf{Ours} methods outperform all baselines on most metrics, especially on the consistency. We conclude that resource-rich neighbors are qualified to provide task priors for low-resource speakers. For our two variants, our full model obtains higher scores on the overall quality and the consistency to the speakers. \textit{Ours$-$SelfEmb}, which skips the training for the target speaker's embeddings, generates more diverse results. This is because absorbing the information from neighbors promotes diversity, and utilizing the embeddings trained on its conversations helps the model to concentrate more on the target speaker's unique characteristics. 

The human evaluation results in Table~\ref{tb:human} are almost consistent with the automatic metrics. The only exception is that \textit{Transformer} has the highest diversity score, but our methods obtain the second highest score in diversity.

\subsection*{Ablation Study}
We demonstrate the performance of the proposed components with an ablation study shown in Table~\ref{tb:ablation}. \textit{$-$NS Loss} shows that removing negative sampling loss from our model causes a performance drop, especially on the appropriateness and consistency. Hence, it is necessary to restrict the task embeddings with the graph structure. \textit{$-$GCN} aggregates the neighbor embeddings by simply averaging it instead of the GCN operation. The performance of replacing the GCN layer also decreases, which verifies the GCN's effect. The model lacking both negative sampling and GCN (\textit{$-$NS Loss $-$GCN}) becomes much worse.
\textit{$-$NS Loss $-$GCN $+$ RAG} means the variant that lacks negative sampling and GCN and follows the retrieval-augmented generation idea \cite{song2018ensemble} to retrieve similar samples from neighbors' conversation samples.
\textit{TA as MAML's base} treats TA as a part of MAML's base model and optimizes TA together with CM instead of optimizing TA across tasks in the outer-loop. Its bad performance suggests that as a cross-task component to capture task embedding and task relation, TA should be optimized across tasks.

\subsection*{Quality of New Speakers' Task Embeddings}
In this section, we examine the quality of new speakers' refined task embedding $\bm{h}_s$ by verifying their ability to distinguish similar speakers from dissimilar ones. As social network neighbors tend to share similar preferences, we regard neighbors to be similar than randomly paired speakers. 
For each new speaker $s$, we calculate two task embedding similarity scores: the similarity of $s$ and one of its neighbor $s_n$, denoted as $sim(s,s_n)$; the similarity of $s$ and a randomly picked speaker $s_r$, denoted as $sim(s,s_r)$. We measure a ratio of $sim(s,s_n)$ being higher than $sim(s,s_r)$ plus a margin $c$, which is defined as $\gamma$-$c$$= \frac{\# (sim(s, s_n) > sim(s, s_r) + c)}{|s|}$.


\begin{table}[htb]
\small
\centering
        \begin{tabular}{c|c|c|c}
        \hline
                               & $\gamma$-0 & $\gamma$-{0.1} & $\gamma$-{0.2} \\ \hline
        Ours (TrainSpeakers)   & 0.99  & 0.99    & 0.89    \\ 
        Ours (RichTestSpeakers) & 0.99  & 0.96    & 0.83    \\ \hline
        Ours    & 0.99  & 0.95    & 0.82    \\ 
        VHUCM+F+SN             & 0.96  & 0.83    & 0.51    \\ 
        Speaker+F+SN              & 0.91  & 0.20    & 0.03    \\ \hline
        \end{tabular}
        \caption{The quality of new speakers' task embeddings. $\gamma$-$c$ is the ratio of new speakers' embeddings achieving higher similarity to neighbors rather than random speaker pairs with a margin $c$.}
        \label{tb:exp31}
    \end{table}

In Table~\ref{tb:exp31}, \textit{Ours (TrainSpeakers)} measures $\gamma$-$c$ on a set of training speakers. \textit{Ours (RichTestSpeakers)} measures $\gamma$-$c$ on new speakers never seen in training, but each speaker can access 60 samples in its support set. \textit{Ours}, \textit{VHUCM+F+SN}, and \textit{Speaker+F+SN} follow the same setting as Sec.~\ref{sec:exp1}, where models measure $\gamma$-$c$ on new speakers never seen in training and access $K$ (i.e. 10) samples in their support sets. We reconstruct the training and testing set and re-train the models, so that we can measure the above five methods on the same set of speakers, which consists of 1,000 speakers.

\begin{figure}[htb]
	\centering
	\includegraphics[width=.46\textwidth]{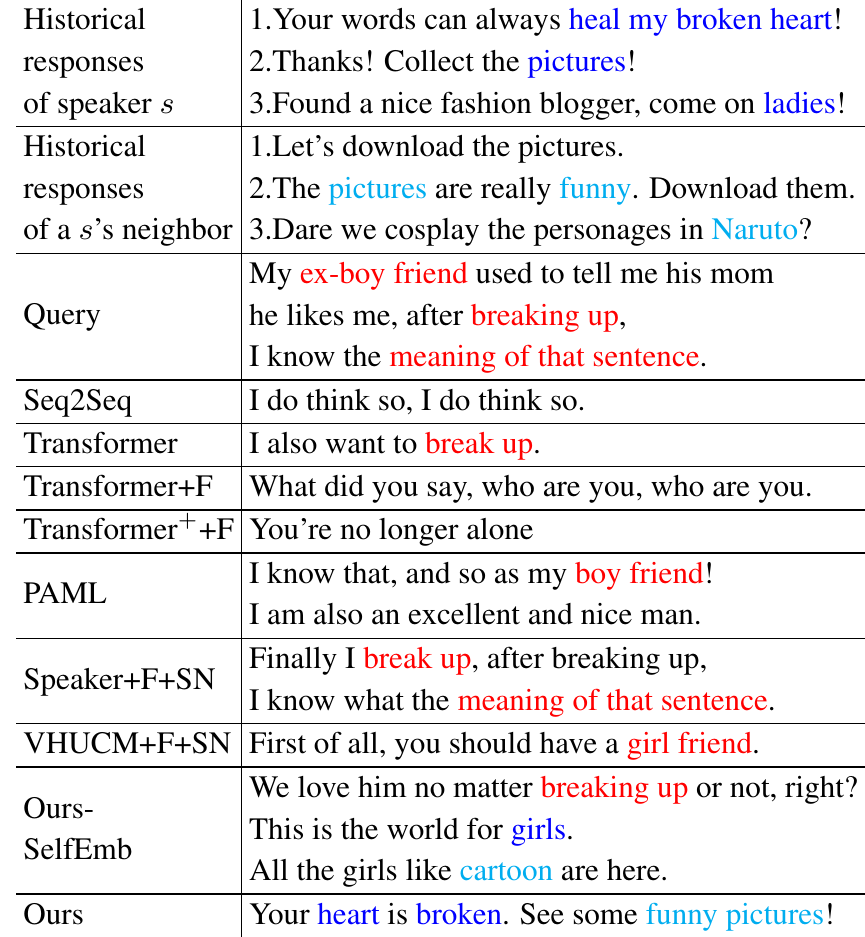}
	\caption{A case with generated responses of all models. We highlight semantic overlaps between generated responses and the query in red color, the responses and speaker's conversation histories in blue, and the responses and neighbors' conversation histories in cyan.}
	\label{fig:casestudy}
\end{figure}

The results in Table~\ref{tb:exp31} show that the quality of new speakers' task embeddings from our model are much higher than that from the baselines.  \textit{Ours (TrainSpeakers)} provides the upper-bound of the task embedding quality since those embeddings are trained by the model, and each speaker has many samples in training. The gap between \textit{Ours (RichTestSpeakers)} and \textit{Ours} is quite small, which reveals that our methods help low-resource speakers to achieve a very close performance to the resource-rich speakers.

\subsection*{Case Study}
Table~\ref{fig:casestudy} shows a case from the test set. In this sample, \textit{Seq2Seq} and \textit{Transformer+F} give general responses. \textit{Transformer} and \textit{Transformer$^+$+F} capture a part of the information in query and make related responses. \textit{Speaker+F+SN} almost repeats the query. \textit{VHUCM+F+SN} outputs a humorous, relevant, and informative response, but it misunderstands the gender of the speaker $s$ ($s$ is probably a girl according to her third historical conversation). \textit{PAML} acts as a girl at the beginning of its responses. As for our methods, \textit{Ours$-$SelfEmb} makes appropriate responses in the view of a girl. Moreover, its response borrows information from the neighbor, where that information (``Naruto" is a ``cartoon") does not appear in the query or the speaker $s$'s conversations. \textit{Ours} can even study some relevant phases from other conversations of the speaker.

\section*{Conclusion}
We propose a few-shot personalized conversation task with a social network, catering for low-resource speakers. In such a scenario, we propose a novel method which enables low-resource speakers to leverage information from the resource-rich speakers with the social network. Our method consists of a PCM and a TA. PCM equips conversational agents with personality via MAML; TA obtains the speaker representation from its few conversations and its resource-rich neighbors in the social network. The representation assists the training of the PCMs. In this way, we remedy the data deficiency issue on low-resource speakers so that our systems are friendly to newcomers or inactive users on social media. The experimental results show our model outperforms all baselines on the overall query, diversity, and consistency.

\section*{Acknowledgments}
Research on this paper was supported by Hong Kong Research Grants Council under grants 16202118 and Tencent AI Lab Rhino-Bird Focused Research Program (No. GF202035).


\bibliography{aaai21}
\bibliographystyle{aaai21}

\end{document}